\documentclass[letterpaper,journal]{IEEEtran}

\usepackage{amsmath,amsfonts,amssymb}
\usepackage{array}
\usepackage[caption=false,font=normalsize,labelfont=sf,textfont=sf]{subfig}
\usepackage{textcomp}
\usepackage{stfloats}
\usepackage{url}
\usepackage{verbatim}
\usepackage{graphicx}
\usepackage{booktabs}
\usepackage{nicefrac}
\usepackage{multirow}
\usepackage{float}
\usepackage{xcolor}
\usepackage{cite}
\usepackage[hidelinks]{hyperref}
\usepackage[capitalise]{cleveref}
\hypersetup{
  pdftitle={Beyond Sequential Interaction: Benchmarking Parallel Execution and Coordination for GUI Agents},
  pdfauthor={Zedong Yu, Qianxing Li, Zhi Gao, Liuyu Xiang, Chenrui Shi, Yang Liu, Huiming Wu, Yujie Wei, Yuhao Fei, Yubo Fu, and Zhaofeng He},
  pdfkeywords={GUI agent, large multimodal model, parallel execution, multi-agent coordination, benchmark}
}
\usepackage{microtype}

\makeatletter
\providecommand{\@LN@col}[1]{}
\providecommand{\@LN@column}[1]{}
\providecommand{\@LN}[2]{}

\makeatother

\newcommand{\etal}{et al.}

\newcommand{\rev}[1]{#1}

\begin{document}

\title{Beyond Sequential Interaction: Benchmarking Parallel Execution and Coordination for GUI Agents}

\author{Zedong Yu\textsuperscript{*}, Qianxing Li\textsuperscript{*},
        Zhi Gao\textsuperscript{\textdagger}, Liuyu Xiang, Chenrui Shi,
        Yang Liu, Huiming Wu, Yujie Wei, Yuhao Fei, Yubo Fu,
        and Zhaofeng He\textsuperscript{\textdagger}%
\thanks{\textsuperscript{*}Zedong Yu and Qianxing Li contributed equally
        to this work.}%
\thanks{Zedong Yu, Liuyu Xiang, Yujie Wei, Yuhao Fei, Yubo Fu, and
        Zhaofeng He are with Beijing University of Posts and
        Telecommunications, Beijing, China (e-mail: yzd0812z2@bupt.edu.cn;
        xiangly@bupt.edu.cn; yujie.wei@bupt.edu.cn; 2022213322@bupt.cn;
        fuyubo23@bupt.edu.cn; zhaofenghe@bupt.edu.cn).}%
\thanks{Zedong Yu, Qianxing Li, Zhi Gao, Chenrui Shi, Yang Liu, and
        Huiming Wu are with the State Key Laboratory for General Artificial
        Intelligence, BIGAI, Beijing, China (e-mail: liuyang1@bigai.ai;
        hwubl@connect.ust.hk).}%
\thanks{Qianxing Li is with China University of Geosciences (Beijing),
        Beijing, China (e-mail: 1004236211@email.cugb.edu.cn).}%
\thanks{Zhi Gao and Chenrui Shi are with Beijing Institute of Technology,
        Beijing, China (e-mail: gaozhibit@bit.edu.cn;
        shichenrui@bit.edu.cn).}%
\thanks{\textsuperscript{\textdagger}Corresponding authors: Zhi Gao and
        Zhaofeng He (e-mail:
        gaozhibit@bit.edu.cn; zhaofenghe@bupt.edu.cn).}%
\thanks{This work has been submitted to the IEEE for possible publication.
        Copyright may be transferred without notice, after which this version
        may no longer be accessible.}}

\markboth{Preprint, July 2026}%
{Yu \MakeLowercase{\textit{et al.}}: Beyond Sequential Interaction}

\maketitle

\begin{abstract}
   Graphical user interface (GUI) agents are systems powered by large multimodal models (LMMs). They perceive screen state and execute user instructions through GUI actions such as clicking, typing, and scrolling on desktops and mobile devices. However, current agents scale poorly to long-horizon tasks: actions incur costly LMM inferences, and performance degrades as context grows. Humans divide such workloads among collaborators who complete sub-tasks in parallel. Yet parallel coordination among GUI agents has received little attention. To close this gap, we introduce ParaGUIBench,
to our knowledge, the first benchmark dedicated to parallel execution and coordination of multiple GUI agents on separate desktop instances. It consists of three components: a multi-device Docker infrastructure with a shared file system; a dataset of 233 tasks spanning six task categories; and an evaluation system with efficiency metrics, including step reduction ratio and token cost. We further introduce ParaGUI, a planner--worker agent that decomposes GUI tasks and dispatches sub-tasks to concurrent workers on separate desktop instances. On ParaGUIBench, ParaGUI reaches a 46.4\% success rate, outperforming the strongest serial baseline (Claude Sonnet 4.6) by 12.9 points while using roughly half the steps and less than half the tokens. \rev{These results show that parallel execution can improve both success rate and efficiency on decomposable, long-horizon GUI tasks, pointing to a direction worth further study.}
Project page: \url{https://github.com/pkgunboat/ParaGUIBench}.
\end{abstract}

\begin{IEEEkeywords}
GUI agent, large multimodal model, parallel execution, multi-agent coordination, benchmark.
\end{IEEEkeywords}

\section{Introduction}
\label{sec:intro}

\IEEEPARstart{G}{raphical} user interface (GUI) agents use Large Multimodal Models (LMMs) to observe the screen and follow user instructions by clicking, typing, and scrolling on computers or phones~\cite{cite1,cite2,cite3,cite4,cite5}. They are part of a broader class of LMM-driven sequential decision systems, where perception, planning, and action are coupled in an interaction loop~\cite{cao2025llmrl,ma2025vlasurvey}. \rev{Most} GUI agents execute long-horizon tasks as a serial chain of perception--decision--action cycles,
with each cycle incurring a costly LMM inference. Abhyankar~\etal~\cite{cite6} report that LMM inference accounts for 75--94\% of per-task latency in GUI agents. Growing context length also makes later steps less accurate and up to 3$\times$ slower than earlier ones.
Model- and system-level acceleration, e.g.\ via quantization, pruning, and distillation~\cite{cheng2026efficient}, lowers per-call latency but leaves this serial loop structure untouched.
Long trajectories thus limit the performance and efficiency of GUI agents.

\begin{figure*}[!t]
    \centering
    \includegraphics[width=1.0\linewidth]{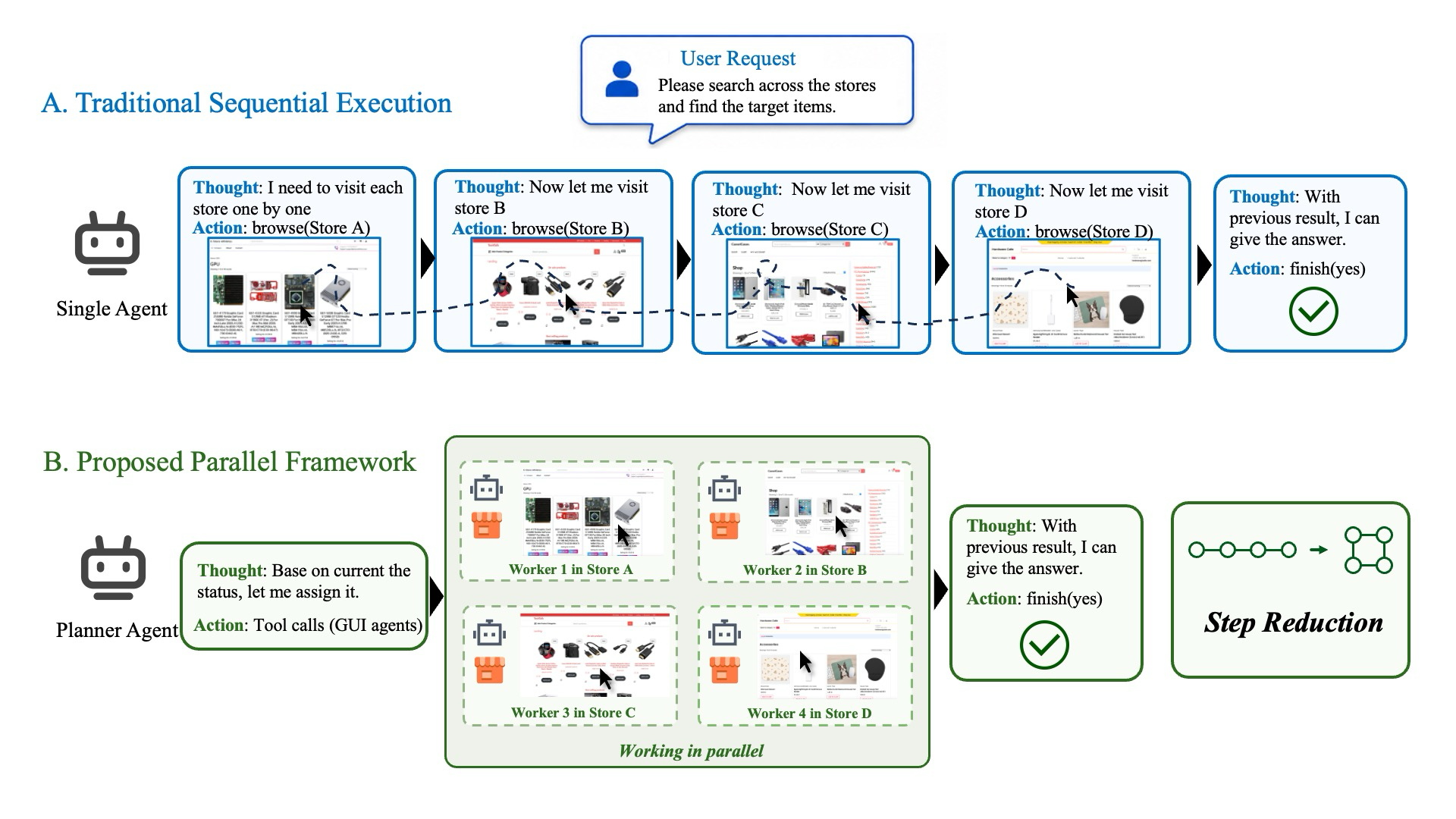}
    \caption{Sequential vs.\ parallel GUI execution. (A) A traditional agent runs a long-horizon task as one serial chain of perception--decision--action steps. (B) ParaGUI's planner decomposes the task and dispatches sub-tasks to parallel GUI workers (four shown; default $N{=}5$), each in its own isolated Docker container, then aggregates their results---shortening the critical path whenever the task can be split.}
    \label{fig:teaser}
\end{figure*}

Humans handle such tasks differently: they divide the workload among teammates, who complete independent sub-tasks in parallel on separate machines before combining their results.
Yet parallel coordination among GUI agents \rev{on separate desktop instances} remains largely unexplored. Standard benchmarks such as OSWorld~\cite{cite7} and WebArena~\cite{cite8} fix evaluation to a single-agent and single-device protocol. \rev{These single-agent, single-device benchmarks} provide no infrastructure for concurrent execution, no tasks that require cross-worker coordination, and no metrics for parallel efficiency. This raises a basic question: \emph{can GUI agents \rev{coordinate} in parallel to accelerate long-horizon GUI workflows, and how do we measure \rev{such coordination?}}

Parallel GUI execution introduces three challenges beyond merely dispatching multiple actions at once. First, the user instruction does not specify the dependencies between steps, so the system must identify which sub-tasks can run concurrently. Second, concurrent workers may conflict when they modify shared artifacts, such as the same file or overlapping output regions. Third, each worker sees only its own sub-task, so local completion does not guarantee that the combined result satisfies the original instruction.

In this paper, we propose ParaGUIBench,
to our knowledge, the first benchmark dedicated to parallel execution and coordination of multiple GUI agents on separate desktop instances. It consists of a parallel-native multi-device Docker infrastructure, a curated dataset, and an evaluation system. The infrastructure runs concurrent workers that share a file system. The dataset contains 233 tasks across six categories grouped under two domains, information retrieval and operation \& manipulation. The evaluation system reports task success rates and three efficiency metrics: step reduction ratio, parallelism degree, and token cost.

We further introduce ParaGUI, a planner--worker agent for parallel GUI automation, as shown in \Cref{fig:teaser}. The planner decomposes a long-horizon GUI task into parallelizable sub-tasks.
It then dispatches them to multiple GUI workers running concurrently on separate desktop instances, aggregates their returned summaries, and decides whether to dispatch another round or terminate. ParaGUI delivers a $+12.9$~percentage-point success-rate gain with roughly half the steps and less than half the tokens of the strongest serial baseline. In addition, \rev{on standard serial OSWorld, ParaGUI outperforms a strong GUI-only baseline, indicating that its planner--worker design transfers beyond parallel-native tasks.}
Using ParaGUI to analyze ParaGUIBench, we find that exploitable parallelism is governed primarily by task dependency structure rather than application domain, while errors within individual GUI workers form the largest failure category.

\noindent\textbf{Contributions.}
(1) We present ParaGUIBench,
to our knowledge the first benchmark dedicated to parallel execution and coordination of multiple GUI agents on separate desktop instances, providing the infrastructure, tasks, and metrics.
(2) We develop ParaGUI, a planner--worker agent for parallel GUI execution that combines adaptive round-based planning with self-contained dispatch---each dispatched instruction contains every earlier result its sub-task depends on, so the worker needs nothing else to run.
(3) We systematically characterize when parallel GUI execution is beneficial, how system components shape capability and efficiency, and why failures occur.

\section{Related Work}
\label{sec:related}

\paragraph{GUI agent benchmarks and efficiency-aware evaluation}
LMM-powered GUI agents have driven a wave of benchmarks. WebArena~\cite{cite8} and VisualWebArena~\cite{cite9} target web tasks; Mind2Web~\cite{cite10} targets offline generalization; OSWorld~\cite{cite7} and WindowsAgentArena~\cite{cite11} target live desktop; AndroidWorld~\cite{cite12} targets mobile. All of them run a single agent on a single device, execute one step at a time, and report the success rate as the main metric. None evaluates coordination, and efficiency is rarely a primary metric. A few works look at cost: Abhyankar~\etal~find current agents are 10--20$\times$ slower than humans~\cite{cite6}, and HAL~\cite{cite13} and WABER~\cite{cite14} add cost and latency to their metrics. These benchmarks that focus on efficiency still assume a single sequential agent rather than multiple agents in parallel.
In contrast, ParaGUIBench evaluates parallel execution and coordination across multiple GUI agents, complementing success rate with step-reduction and parallelism metrics that these single-agent benchmarks omit.

\paragraph{Parallel and multi-agent systems}
Prior work explores parallel execution for LLM agents at the API level (such as LLMCompiler~\cite{cite15} and DynTaskMAS~\cite{cite16}) and at the planning level (such as Plan-over-Graph~\cite{cite17} and Flow~\cite{cite33}). Frameworks such as MetaGPT~\cite{cite18} and MegaAgent~\cite{cite19} organize agents into role-based workflows. A separate multi-agent reinforcement learning literature studies coordination through communication and task allocation~\cite{wu2026gcm,hu2025routing}, but mainly in learned-control settings rather than benchmarked GUI workflows.
\rev{The closest adjacent efforts are UFO$^3$~\cite{cite20} and Flash-Searcher~\cite{cite21}.} UFO$^3$~\cite{cite20} evaluates asynchronous \rev{coordination} across heterogeneous Windows and Linux endpoints, where parallelism is organized primarily across device agents and its Windows endpoint itself combines GUI and application-API control. Flash-Searcher~\cite{cite21} parallelizes web information retrieval through search and crawl APIs and reports execution-step efficiency, but involves no screen-grounded interaction or GUI-state modification. ParaGUIBench instead directly evaluates multiple GUI agents operating concurrently on separate desktop instances, \rev{with a rule-based evaluator that checks each task outcome against} textual answers or the resulting environment state.
On the evaluation side, PLaG~\cite{cite22}, Robotouille~\cite{cite23}, AsyncTool~\cite{cite24}, MultiAgentBench~\cite{cite25}, and Collab-Overcooked~\cite{cite26} test asynchronous planning and coordination in \emph{non-GUI, abstracted environments}---text-based games, symbolic task graphs, and gridworld-style kitchens---where actions are high-level API calls rather than pixel-level GUI operations.
Different from them, ParaGUIBench grounds parallel coordination in \emph{real, stateful GUI environments}, where agents act through screen-grounded controls and may contend over persistent shared artifacts. \Cref{tab:related} contrasts these systems and benchmarks by their \rev{primary target, parallel} unit, interaction interface, and evaluation emphasis.

\begin{table*}[!t]
  \caption{Comparison of representative systems and benchmarks for parallel agent execution.\label{tab:related}}
  \centering
  \footnotesize
  \setlength{\tabcolsep}{4pt}
  \renewcommand{\arraystretch}{1.2}
  \begin{tabular}{@{}>{\raggedright\arraybackslash}p{2.7cm}>{\raggedright\arraybackslash}p{2.6cm}>{\raggedright\arraybackslash}p{2.9cm}>{\raggedright\arraybackslash}p{2.6cm}>{\raggedright\arraybackslash}p{3.7cm}@{}}
    \toprule
    Work & Primary target & Parallel unit & Interaction interface & Evaluation emphasis \\
    \midrule
    OSWorld~\cite{cite7} & Single-agent GUI control & None (single serial agent) & Screen, mouse, keyboard & Task success \\
    MultiAgentBench~\cite{cite25} & \rev{Multi-agent coordination} & LLM agents & Text / API calls & Success, coordination \\
    Collab-Overcooked~\cite{cite26} & Cooperative planning & Agents in a gridworld & Symbolic actions & Success, coordination \\
    Flash-Searcher~\cite{cite21} & Parallel web retrieval & DAG reasoning / tool branches & Search and crawl APIs & Accuracy, execution steps \\
    UFO$^3$ / NebulaBench~\cite{cite20} & \rev{Cross-device coordination} & Heterogeneous device agents & CLI + hybrid GUI/API & Success, parallelism, latency, robustness \\
    \midrule
    \textbf{ParaGUI / ParaGUIBench (ours)} & Parallel GUI execution & Multiple GUI workers & Screen, mouse, keyboard & Success, critical-path steps, parallelism, tokens \\
    \bottomrule
  \end{tabular}
\end{table*}

\section{ParaGUIBench}
\label{sec:framework}

ParaGUIBench consists of three components: a parallel-native multi-device infrastructure with \rev{a shared file system}, a curated dataset of 233 tasks \rev{spanning six categories under two domains (each also annotated by its parallel-execution pattern)}, and an evaluation system that jointly measures task success and parallel efficiency.

\subsection{Parallel-Native Infrastructure}
\label{sec:infrastructure}

To isolate GUI control while preserving file-level coordination, each worker runs in a separate Docker container with its own full Ubuntu desktop, while all containers mount a shared directory for exchanging intermediate files. Separate desktops give each worker its own mouse cursor and keyboard focus, so concurrent workers cannot disrupt one another's clicks and keystrokes, whereas the shared directory preserves the artifacts needed for coordination. \rev{The infrastructure} runs up to $N$ parallel workers ($N{=}5$ by default).
A multi-container execution manager replaces OSWorld's \rev{single virtual-machine} backend. The infrastructure supports a configurable number of workers and configurable per-container resources, and restores every container to a clean state before each run.

\subsection{Task Construction}
\label{sec:tasks}

We organize task construction into four parts: defining task categories, annotating parallel-execution patterns, collecting tasks from complementary sources, and applying quality-control filters.

\subsubsection{Task Categories}
\label{sec:taskcat}

The tasks span six task categories grouped under two application domains, summarized in \Cref{tab:task_stats}.
We choose these categories to cover both \rev{information retrieval and operation \& manipulation} across browsers, office applications, and the file system.

\emph{Information retrieval} (77 tasks) asks agents to navigate digital environments and produce a textual answer. We verify each answer by text-based matching. It covers web search \rev{in} the browser
and file search \rev{over Word, PowerPoint, and other document} files.

\emph{Operation \& manipulation} (156 tasks) requires \rev{the agent to directly modify} the environment state. We verify whether the task is completed by inspecting the application configurations or the file system. The operation \& manipulation domain covers online shopping,
file operation,
web navigate,
and search \& write tasks. 
\Cref{tab:task_stats} \rev{lists each task category, its internal sub-categories, and one representative task instruction, illustrating} the diversity of GUI interactions in ParaGUIBench.

\begin{table*}[!t]
    \caption{Task dataset statistics with representative examples. \# means the number of tasks.\label{tab:task_stats}}
    \centering
    \small
    \setlength{\tabcolsep}{4pt}
    \renewcommand{\arraystretch}{1.0}
    \begin{tabular}{@{}lp{0.32\linewidth}rp{0.48\linewidth}@{}}
      \toprule
      \textbf{Domain} & \rev{\textbf{Category}} & \textbf{\#} & \textbf{Example Task Instruction}
\\
      \midrule
      \multirow{2}{*}{\shortstack[l]{Info.\\Retrieval}}
        & web search (Statistic / Visual / Conditional / Multi-hop) & 65
        & \textit{How many Science magazines were published in 2024 that feature a fish on their cover?} \\
      \cmidrule(lr){2-4}
        & file search (PPT / Word / generic) & 12
        & \textit{There are some files in the folder. Which AI-related papers have cartoon-style
illustrations?} \\
      \midrule
      \multirow{4}{*}{\shortstack[l]{Oper.\\Manip.}}
        & online shopping --- WebMall (11 sub-categories) & 91
        & \textit{Find the cheapest offer for the Samsung Galaxy S24 Plus. If multiple offers share the
lowest price, return all of them.} \\
      \cmidrule(lr){2-4}
        & file operation (Combination / Batch (Word/Excel/PPT) / Coding) & 42
        & \textit{The folder contains many Word, PPT, and Excel files. Create different subfolders based on
themes and content, and classify them into the subfolders.} \\
      \cmidrule(lr){2-4}
        & web navigate (Bookmark / Settings) & 13
        & \textit{Navigate to Tesla's official website, open the pages for Model Y, Model 3, and Model S
for comparison, and add each page to the bookmarks.} \\
      \cmidrule(lr){2-4}
        & search \& write (File / Web) & 10
        & \textit{Please help me fill in the top five schools in the 2025 QS rankings and their detailed
information into this table.} \\
      \midrule
      \multicolumn{2}{@{}l}{\textbf{Total}} & \textbf{233} & \\
      \bottomrule
    \end{tabular}
\end{table*}

\subsubsection{\rev{Parallel-Pattern Classes}}
\label{sec:taskpattern}

We characterize each task by the dependency structure of the trajectory that solves it, using three classes that we carry through to our results: \textit{parallel\_independent}, \textit{parallel\_dependent}, and \textit{serial}. \textit{parallel\_independent} tasks decompose into sub-tasks with no inter-dependency: workers share no state and need no cross-worker communication, so all sub-tasks run concurrently (e.g., querying several independent stores, or assigning one file per worker). \textit{parallel\_dependent} tasks decompose into a directed acyclic graph (DAG) of sub-tasks: some sub-tasks consume the outputs of earlier ones, yet independent branches still run in parallel (e.g., a multi-hop query that splits into parallel sub-queries and then merges them). \textit{serial} tasks admit no exploitable parallelism; \rev{because the dependency chain is unavoidable, the planner concentrates dispatch on essentially a single worker, driving the parallelism degree down toward~1.}

\subsubsection{Task Sources}
\label{sec:tasksources}

We combine two task sources for complementary purposes. Existing benchmarks provide realistic applications, stable environments, and reusable evaluation infrastructure, whereas manually constructed tasks fill gaps in long-horizon decomposition and cross-worker coordination. Specifically, we adapt existing benchmarks and environments as the first source:
From VeriWeb~\cite{cite28}, we collect \rev{information-retrieval} tasks with deterministic answers.
From OSWorld~\cite{cite7}, we \rev{combine several simple tasks into a single long-horizon task that decomposes into parallel sub-tasks.}
From WebMall~\cite{cite27}, we reformulate \rev{transient-state tasks into checkpoint-verifiable ones (e.g., ``reach a fixed product page'' becomes ``bookmark the target URL''); the verification protocol is detailed in \Cref{sec:evaluation}.}
The second source is manually constructed tasks for parallel execution and cross-worker coordination. They span each task category except online shopping, which is built entirely from the WebMall environment.

\subsubsection{Quality Control}
\label{sec:qc}

Reliable automated evaluation requires each task to have a unique, stable target and to demand genuine GUI interaction. We therefore apply two filters. The uniqueness filter rewrites or removes tasks with multiple valid answers or unstable targets. The shortcut filter removes tasks that an LMM can solve without \rev{genuine GUI interaction;} human annotators validate candidates and inspect trial-run logs before inclusion.

\subsection{Evaluation System}
\label{sec:evaluation}

ParaGUIBench measures task success rate and agent efficiency. \rev{Efficiency comprises three metrics: the step reduction ratio, the parallelism degree, and token cost.}
\paragraph{Success Rate (SR)}
SR is the fraction of tasks solved correctly under the verification protocol.

\paragraph{Step Reduction Ratio ($S$)}
The step reduction ratio measures how much parallel execution shortens the longest chain of sequential steps. Let $\tau_{\mathrm{serial}} = \{(o_t, a_t)\}_{t=1}^T$ denote the trajectory of a single-worker GUI-only agent that runs the same task sequentially. The baseline length is
\[ L_{\mathrm{serial}} = |\tau_{\mathrm{serial}}| = T, \]
where $T$ is the number of perception--decision--action cycles executed by this agent. This trajectory is regarded as the serial baseline. Since we report $S$ against more than one such baseline agent (the worker-equivalent Seed-1.8 and the strongest serial agent, Claude Sonnet 4.6), each table states which baseline its $S$ is computed against. A parallel run executes in $R$ sequential planner rounds. Round $r$ dispatches $K_r$ workers. Worker $k$ in round $r$ runs $s_{r,k}$ steps. We define the parallel length $L_{\mathrm{parallel}}$ and the step reduction ratio $S$ as
\begin{equation}
\begin{aligned}
L_{\mathrm{parallel}} &= \sum_{r=1}^{R} \max_{k \in \{1, \ldots, K_r\}} s_{r,k}, \\
S &\triangleq \frac{L_{\mathrm{serial}}}{L_{\mathrm{parallel}}}.
\end{aligned}
\end{equation}
\rev{Here $S{>}1$ means parallel execution uses fewer critical-path steps than the serial baseline, while $S{<}1$ means it uses more.}
Within each round, workers execute concurrently, so the round length is the largest per-worker step count, $\max_{k} s_{r,k}$. Summing these round lengths gives $L_{\mathrm{parallel}}$, which we call the \emph{critical-path steps}\rev{; it counts only worker GUI steps and excludes planner reasoning and aggregation calls}. This is the \emph{steps} column reported in all result tables.
$S$ contrasts \emph{two different agent systems} (a single-agent GUI-only baseline vs.\ the planner--worker ParaGUI) on the same task.
We report $S$ at the corpus level, as the ratio of the mean $L_{\mathrm{serial}}$ to the mean $L_{\mathrm{parallel}}$ over all tasks---a task-weighted aggregate consistent with the steps in all result tables and with the pooling used for $P$.

\paragraph{Parallelism ($P$)}
The parallelism degree quantifies how effectively the planner decomposes a task.
Let the total number of worker steps actually expended be
\[ L_{\mathrm{total}} = \sum_{r=1}^{R} \sum_{k=1}^{K_r} s_{r,k}. \]
We define $P \triangleq L_{\mathrm{total}} / L_{\mathrm{parallel}}$. A value of $P{=}1.0$ indicates fully serialized dispatch, while $P{=}N$ indicates ideal $N$-way parallelism, upper-bounded by the per-task worker count. Unless stated otherwise, the per-task step counts we report for ParaGUI are \emph{critical-path steps} ($L_{\mathrm{parallel}}$), not total worker steps ($L_{\mathrm{total}}$).

\paragraph{Token Cost}
We sum the token cost of the planner and all workers. This captures the resource overhead from parallelism. We count every token in full, with no prompt-cache savings.

\paragraph{Evaluation Procedures}
We tailor the verification protocol to each task type and read the verdict from the real environment state rather than the agent's self-report. We mark a task solved only when the agent recovers all expected items with nothing extraneous. We check information-retrieval answers by normalization and fuzzy matching: punctuation/format normalization, keyword coverage, numeric tolerance, and multi-value F1. This lets correct answers in different surface forms still pass. For web navigate and search-style online shopping tasks, verifying transient page state is hard. We sidestep this by reformulating ``reach page X'' as ``bookmark page X''. The agent saves the target page(s); the evaluator reads the browser's bookmark file and matches saved URLs against the expected targets. For cart and checkout shopping tasks, we verify from the live accessibility tree (the shopping cart and the order-confirmation page). We inspect file-operation outputs directly (cell values, formatting, document structure). We score search \& write tasks cell-by-cell against a gold spreadsheet.

\section{ParaGUI}
\label{sec:paragui}
\begin{figure*}[!t]
  \centering
  \includegraphics[width=1.0\linewidth]{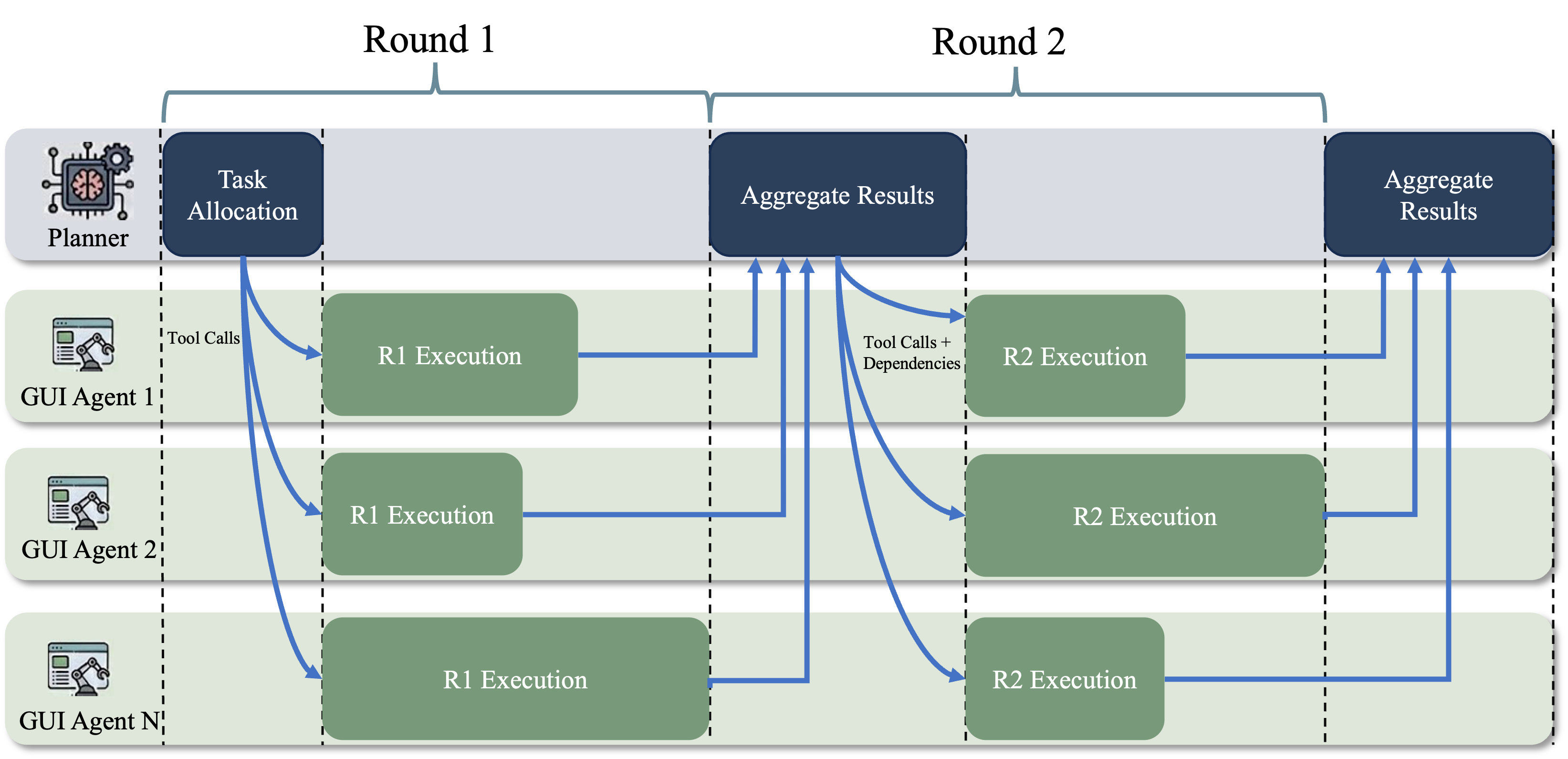}
  \caption{The ParaGUI planner--worker architecture. Each round, the planner dispatches a batch of sub-tasks to up to $N$ GUI workers running in parallel, aggregates their returned summaries, and decides whether to dispatch another round or terminate.
  }
  \label{fig:paragui}
\end{figure*}

\Cref{fig:paragui} shows the planner--worker architecture of ParaGUI. ParaGUI adapts this pattern to parallel GUI execution through three design choices: adaptive round-based planning, self-contained dispatch, and \rev{disjoint-region assignment}. Together, these let the planner revise later dispatches from intermediate outputs, write into each instruction the earlier results its sub-task depends on, and reduce conflicts over shared artifacts. Each worker executes its assigned instruction in an isolated Docker container.

\subsection{Round-based Execution Workflow}
\label{sec:agent_workflow}

ParaGUI's planner \rev{exploits the parallel tool-calling of modern LMMs: in a single inference step it emits a batch of parallel invocations, each dispatching one} sub-task to one worker. ParaGUI runs a round-based loop. The planner alternates between \emph{\rev{reasoning}} steps and synchronous parallel worker dispatch. The loop starts with an empty interaction history $H$ and round counter $r{=}0$. In each round $r$, the planner conditions on the user instruction and the accumulated history $H$, then emits a batch of parallel invocations. Each invocation specifies the target worker index $i$, a natural-language sub-task $g_i$, and a set of dependency references $D_i$ that point to the outputs of earlier rounds.

Rather than chaining sub-tasks into a sequence, ParaGUI emits self-contained sub-tasks that can be dispatched concurrently.
We call this dispatch protocol \emph{self-contained dispatch} (SCD). Each worker can see only the single instruction it is handed---not the planner's reasoning, and not what the other workers did---so any result a sub-task needs from an earlier round must be written into that instruction before the worker starts. Each invocation targets one selected worker and carries a natural-language sub-task together with a set of symbolic references to the outputs of earlier rounds. The planner marks each reference inline in its thought with a lightweight tag (e.g.\ \texttt{<dep round="3" agent="2"/>}); a regex parses these tags into the dependency set $D_i$, and when parsing fails, a heuristic analyzer reconstructs $D_i$ from the wording of the sub-task. Before dispatch, the planner resolves every reference to a concrete string or file path and splices it into the sub-task, turning it into a self-contained instruction with no unresolved references. The worker then executes it on its own, never reading the planner's history or any other worker's state.

All invocations in a round execute concurrently, one per Docker container. The planner blocks until every worker returns. Each worker returns a textual summary along with the paths of any files it produced. The planner does \emph{not} receive worker screenshots, which keeps its context length bounded as $N$ scales. We append these summaries to $H$ and start the next round for the planner.

After each round, ParaGUI checks two termination conditions. First, the planner can emit a \texttt{<answer>\allowbreak...\allowbreak</answer>} tag in its thought; its payload becomes the task output. Second, the \rev{global GUI-step budget runs out, after which no further worker dispatch is allowed and a synthetic ``budget exhausted'' message forces the planner to commit a final answer in the next round. In either} case, the evaluation protocol scores the final environment state regardless of whether the agent produced an explicit answer.

\subsection{Planner}
\label{sec:planner}

Each round, the planner \rev{decides the next dispatch, mitigates conflicts over shared artifacts, and aggregates the} returned results. First, it identifies sub-tasks that are independent and safe to dispatch concurrently, while deferring those that depend on the current round's outputs. Unlike static DAG planners~\cite{cite15,cite17}, which fix the dependency graph upfront, ParaGUI revises later dispatches from intermediate worker summaries. The planner may also re-target the same worker when preserving its UI state is useful. Second, when multiple sub-tasks touch a shared artifact, the planner assigns non-overlapping responsibilities (\emph{disjoint-region assignment}) where possible and serializes operations that genuinely conflict; these decisions come from the planner LMM over the accumulated history $H$, with no hand-coded resource manager. Third, after all workers return, the planner aggregates their textual summaries and produced file paths, updates the interaction history, and decides whether to dispatch another round or terminate.

\subsection{Worker}
\label{sec:worker}

ParaGUI treats each worker as an interchangeable GUI executor rather than introducing a specialized worker policy. Each worker controls a single Docker container and shares the mounted file system with the others. Workers reuse the OSWorld primitive action space~\cite{cite7} (click, scroll, type, hotkey, $\dots$). This lets us swap in any vision-capable LMM as a worker backbone. On invocation, a worker receives the natural-language sub-task $g_i$ with the dependency references $D_i$ already resolved to concrete strings or file paths. The worker then returns a structured summary: a textual report of its actions and the paths of any files produced. Cross-worker communication uses only the \rev{shared directory} \texttt{/home/user/shared/}. We do not support direct messaging between workers.

\paragraph{Step budgets and explicit failure signals}
Each worker invocation runs at most 25 perception--decision--action steps. The global GUI-step budget across all workers in one task is 200 steps. ParaGUI propagates two explicit failure signals to the planner: \rev{(i) a \emph{step-cap overflow}, when a worker reaches its per-invocation step cap and returns \texttt{FAIL}; and (ii) a \emph{voluntary failure}, when} a worker invokes \texttt{fail()} on an unrecoverable UI state. The per-invocation cap of 25 steps balances sub-task completion against token cost: a larger budget lets a stuck worker accumulate redundant \rev{actions, and thus spend more} tokens, while a smaller budget leaves too little room to complete a non-trivial sub-task. \rev{By contrast,} hitting the global cap injects a synthetic ``budget exhausted'' message that forces the planner to commit a final answer in the next round.

\section{Experiments and Analysis}
\label{sec:experiments}

We organize our experiments around five research questions. \textbf{RQ1:} How does ParaGUI compare with serial GUI agents in task success and efficiency? \textbf{RQ2:} Under what task conditions is parallel GUI execution beneficial? \textbf{RQ3:} What failure modes \rev{arise when ParaGUI runs on ParaGUIBench?} \textbf{RQ4:} Which components and configurations govern ParaGUI's capability--efficiency trade-off? \textbf{RQ5:} Does the planner--worker architecture transfer to standard serial GUI tasks? RQ1 reports the full 233-task comparison together with \rev{a paired analysis on the \emph{agree-subset}---tasks on which ParaGUI and a baseline have the same outcome (both correct or both fail);} RQ2 and RQ3 analyze the full ParaGUIBench; RQ4 uses a 51-task diagnostic subset; and RQ5 evaluates the full 369-task OSWorld suite with $N{=}1$.

\subsection{Experimental Setup}
\label{sec:setup}

\paragraph{Models}
We evaluate a set of representative LMMs across the planner--worker architecture. ParaGUI uses GPT-5.4~\cite{cite30} as its planner backbone and Doubao Seed-1.8~\cite{cite31} (Seed-1.8 hereafter) as its worker backbone. This pairing combines strong long-context reasoning on the planner side with broad action-space coverage and lower per-step cost on the worker side. As serial GUI-only baselines, we evaluate four single GUI agents (no planner), each driven by one LMM backbone and competitive among publicly reported GUI agents on OSWorld and related benchmarks at submission time. The four are Seed-1.8, Qwen3-VL-235B-A22B-Instruct~\cite{qwen3vl}, Holo3-35B-A3B~\cite{holo3}, and Claude Sonnet 4.6~\cite{claude46}. Together they span general-purpose and GUI-specialized model families.

\paragraph{Execution Configurations}
We evaluate two kinds of systems. The four serial baselines run in \textit{GUI-only} mode: a single GUI agent executing one perception--decision--action loop per step, with no planner.
ParaGUI runs in \textit{Parallel} mode with $N{=}5$ workers concurrently on separate isolated Docker containers, each allocated 4\,GiB of RAM and 2 vCPUs. \rev{Each serial GUI-only baseline runs a single agent in one such container under the same per-container resources. The} planner coordinates the workers and does not occupy a Docker container. Ablations also vary the planner and worker backbones and the parallelism degree $N$. Unless otherwise noted, every agent is decoded with temperature $0$ and a fixed random seed of $42$ for reproducibility.

\paragraph{Visual-History Configuration}
Each agent runs under its officially recommended visual-history retention policy: Seed-1.8 keeps the most recent 3 screenshots, Claude Sonnet 4.6 retains up to 10 with adaptive sizing, and Qwen3-VL \rev{and Holo3 each keep the most recent 5 screenshots.} ParaGUI workers (Seed-1.8 backbone) inherit Seed-1.8's 3-screenshot policy. The GPT-5.4 planner consumes text-only worker summaries and receives no screenshots, so its token consumption does not scale with worker visual context. Token gaps reported in \Cref{tab:main_results,tab:efficiency_partition,tab:efficiency_partition_claude} therefore reflect both intrinsic per-call cost and these visual-history configuration differences. We report numbers under each model's actual deployment configuration rather than normalize across agents, because (a) official configurations are how each model is actually deployed in practice and (b) any normalization would itself introduce a hidden axis of comparison. An additional ablation restricts Claude Sonnet 4.6 to a 3-screenshot budget and shows that the token gap does \emph{not} narrow under this normalization---in fact, the restricted Claude configuration spends more steps and more \rev{tokens at a slightly lower success rate (full numbers in \Cref{tab:ablation_full}).}

\paragraph{Evaluation Split}
Vision-intensive multi-agent runs are expensive, so we adopt a two-tier evaluation. Our main evaluation reports each method on the full 233-task benchmark under its default configuration. For sweeping many planner and worker combinations and fine-grained ablation studies (e.g.\ planner vs.\ worker impact), we use a 51-task subset (file search 7, web search 10, online shopping 20, web navigate 2, file operation 9, search \& write 3). This subset is selected to approximate the benchmark's domain mix while keeping compute tractable.

\subsection{RQ1: Serial-Agent Comparison}
\label{sec:main_results}

On the full 233-task benchmark, ParaGUI achieves a higher success rate than the strongest serial baseline (46.4\% vs.\ 33.5\%) while using roughly half as many critical-path steps and fewer than half as many tokens. The full-benchmark comparison establishes these success-rate gains; at matched outcomes, the paired analyses then show that ParaGUI uses more tokens and does not consistently reduce steps relative to Seed-1.8, but uses fewer steps and tokens than Claude Sonnet 4.6.

We compare ParaGUI (Planner+Worker, $N{=}5$) against four representative serial GUI-only baselines on the full ParaGUIBench (233 tasks; baselines introduced in \Cref{sec:setup}). \Cref{tab:main_results} reports \rev{category-level} SR, critical-path steps, and tokens, grouped by domain.

\begin{table*}[!t]
    \caption{Main results on ParaGUIBench (233 tasks). ParaGUI ($N{=}5$) vs.\ four serial GUI-only baselines. SR: Success Rate; steps: avg.\ critical-path steps; Tokens: avg.\ per-task token usage (M). \rev{$\uparrow$/$\downarrow$: higher/lower is better.}
    \label{tab:main_results}}
    \centering
    \resizebox{\linewidth}{!}{%
    \begin{tabular}{@{}llr*{15}{c}@{}}
      \toprule
      & & & \multicolumn{5}{c}{\textbf{SR \% (avg)}~\rev{$\uparrow$}} & \multicolumn{5}{c}{\textbf{Critical path (avg)}~\rev{$\downarrow$}} &
  \multicolumn{5}{c}{\textbf{Tokens (avg, M)}~\rev{$\downarrow$}} \\
      \cmidrule(lr){4-8} \cmidrule(lr){9-13} \cmidrule(lr){14-18}
      \textbf{Domain} & \rev{\textbf{Category}} & \textbf{\#}
        & \textbf{ParaGUI} & \textbf{Seed-1.8} & \textbf{Qwen3-VL} & \textbf{Holo3} & \textbf{Claude
  4.6}
        & \textbf{ParaGUI} & \textbf{Seed-1.8} & \textbf{Qwen3-VL} & \textbf{Holo3} & \textbf{Claude
  4.6}
        & \textbf{ParaGUI} & \textbf{Seed-1.8} & \textbf{Qwen3-VL} & \textbf{Holo3} & \textbf{Claude
  4.6} \\
      \midrule
      \multirow{2}{*}{Info.\ Retrieval}
        & web search      &  65 &  \textbf{41.5} &  24.6 & 12.3 &  6.1 & 32.3 & 38.9 & 35.9 & 47.4 & 125.1 & 87.3
  & 1.07 & 0.38 & 0.58 & 3.15 & 2.48 \\
        & file search     &  12 &  \textbf{75.0} &   8.3 & 16.7 & 33.3 & 33.3 & 20.4 & 12.6 & 36.1 &  44.5 & 74.7
  & 0.23 & 0.10 & 0.41 & 0.90 & 0.45 \\
      \midrule
      \multirow{4}{*}{Operation}
        & online shopping &  91 &  \textbf{54.9} &  38.5 &  9.9 &  2.2 & 44.0 & 32.7 & 37.8 & 48.0 & 148.9 & 77.6
  & 0.69 & 0.38 & 0.59 & 3.65 & 2.17 \\
        & file operation  &  42 &  \textbf{33.3} &  21.4 & 11.9 & 14.3 & 21.4 & 54.5 & 44.1 & 64.8 & 108.3 & 72.8
  & 0.95 & 0.49 & 0.84 & 3.06 & 2.09 \\
        & web navigate    &  13 &  \textbf{53.8} &  23.1 & 15.4 &  0.0 & 23.1 & 29.2 & 28.8 & 72.9 & 149.6 & 39.5
  & 0.44 & 0.28 & 0.93 & 3.19 & 0.66 \\
        & search \& write &  10 &  10.0 &  10.0 &  0.0 &  0.0 & 10.0 & 59.2 & 39.8 & 75.3 &  81.2 & 48.0
  & 0.87 & 0.46 & 1.01 & 2.20 & 1.22 \\
      \midrule
      \multicolumn{2}{@{}l}{\textbf{Total}} & 233
        & \textbf{46.4} & 27.9 & 11.2 & 6.9 & 33.5
        & 38.7 & 36.7 & 52.8 & 126.7 & 75.9
        & 0.81 & 0.38 & 0.66 & 3.18 & 2.03 \\
      \bottomrule
    \end{tabular}}%
\end{table*}

\paragraph{Information Retrieval}
On the information-retrieval domain (\Cref{tab:main_results}, 77 tasks), ParaGUI outperforms the strongest serial baseline Claude Sonnet 4.6 on both \rev{categories}: web search ($41.5\%$ vs.\ $32.3\%$) and file search ($75.0\%$ vs.\ $33.3\%$). It does so with far fewer critical-path steps (web search $38.9$ vs.\ $87.3$; file search $20.4$ vs.\ $74.7$) and fewer tokens (web search $1.07$M vs.\ $2.48$M; file search $0.23$M vs.\ $0.45$M).

\paragraph{Operation \& Manipulation}
On the operation domain, ParaGUI outperforms Claude Sonnet 4.6 \rev{on the three categories whose sub-tasks can run on independent workers}: online shopping (54.9\% vs.\ 44.0\%, +10.9~points), file operation (33.3\% vs.\ 21.4\%, +11.9~points), and web navigate (53.8\% vs.\ 23.1\%, +30.7~points). Search \& write\rev{, whose sub-tasks must edit one shared document,} remains the hardest category for both systems (10.0\% vs.\ 10.0\%).

\paragraph{Efficiency and Cost}
Against the strongest serial baseline Claude Sonnet 4.6, ParaGUI attains higher SR (46.4\% vs.\ 33.5\%, +12.9~points). It uses roughly half the critical-path steps (38.7 vs.\ 75.9) and less than half the tokens (0.81M vs.\ 2.03M) under each model's default visual-history configuration.
Against Seed-1.8---a token-efficient baseline that is also our worker backbone---ParaGUI uses comparable steps (38.7 vs.\ 36.7) and delivers a +18.5-point SR gain (46.4\% vs.\ 27.9\%). ParaGUI spends about $2.1\times$ \rev{as many tokens as} Seed-1.8 (0.81M vs.\ 0.38M); this overhead is partly explained by a baseline behavior pattern, in which Seed-1.8 frequently terminates early on incorrect trajectories with high confidence, well before the step budget (\Cref{tab:fail_step_dist}), lowering its step and token counts without improving its success rate. Because Seed-1.8 is the worker in both systems, the +18.5-point gain cannot be attributed to a stronger worker backbone; it must arise from the planner--worker architecture, including decomposition, SCD, and parallel dispatch.
To separate efficiency from accuracy, we next restrict the comparison to the \emph{agree-subset}: tasks on which ParaGUI and the baseline have the same outcome (both correct or both fail). We focus this analysis on Seed-1.8 ($170/233$ paired tasks; \Cref{tab:efficiency_partition})---the token-efficient worker backbone---since ParaGUI already outperforms Claude Sonnet 4.6 on both success rate and efficiency (\Cref{tab:efficiency_partition_claude}); the complementary Claude Sonnet 4.6 comparison appears there.

\Cref{tab:efficiency_partition} reports a breakdown of the agree-subset by partition (both-correct vs.\ both-fail) and category. To keep each row well populated, it shows only categories with $\ge 10$ paired tasks; the remaining 31 paired tasks fall in categories with fewer than 10 each.

\begin{table}[!t]
    \caption{Efficiency on the paired agree-subset (170/233 tasks): ParaGUI ($N{=}5$) vs.\ the serial Seed-1.8 baseline (its worker backbone), for categories with ${\geq}10$ paired tasks. Using the same Seed-1.8 GUI executor controls for worker-backbone strength and isolates the effect of the planner--worker architecture. \emph{Step Red.}: baseline steps $\div$ ParaGUI steps ($\uparrow$, ${>}1$ means ParaGUI uses fewer steps). \emph{Token}$\times$: ParaGUI tokens $\div$ baseline tokens ($\downarrow$, ${<}1$ means ParaGUI is cheaper). \rev{The critical-path column lists ParaGUI and Seed-1.8 mean steps side by side, not a ratio.}\label{tab:efficiency_partition}}
    \centering
    \small
    \resizebox{\columnwidth}{!}{%
    \begin{tabular}{@{}llrcc c@{}}
      \toprule
      \textbf{Partition} & \rev{\textbf{Category}} & \textbf{\#} & \rev{\textbf{Critical path (Para vs.\ Seed)}} & \textbf{Step Red.\ ($S$)}~$\uparrow$ & \textbf{Token$\times$}~$\downarrow$ \\
      \midrule
      \multirow{2}{*}{Both Correct}
        & online shopping & 29 & 31.7\,/\,43.1 & \textbf{1.36$\times$} & 1.12$\times$ \\
        & web search      & 13 & 31.1\,/\,25.6 & 0.82$\times$          & 4.35$\times$ \\
      \midrule
      \multirow{3}{*}{Both Fail}
        & online shopping & 35 & 38.8\,/\,37.1 & 0.96$\times$ & 2.46$\times$ \\
        & web search      & 34 & 43.1\,/\,41.3 & 0.96$\times$ & 2.62$\times$ \\
        & file operation  & 28 & 52.9\,/\,43.1 & 0.82$\times$ & 2.04$\times$ \\
      \bottomrule
    \end{tabular}}%
\end{table}

\begin{table}[!t]
    \caption{Efficiency on the paired agree-subset vs.\ the Claude Sonnet 4.6 baseline, for categories with ${\geq}10$ paired tasks; complementary to \Cref{tab:efficiency_partition}. \emph{Step Red.}: baseline steps $\div$ ParaGUI steps ($\uparrow$, ${>}1$ means ParaGUI uses fewer steps). \emph{Token}$\times$: ParaGUI tokens $\div$ baseline tokens ($\downarrow$, ${<}1$ means ParaGUI is cheaper). Rows shown cover 129 paired tasks; the remaining paired tasks fall in categories with fewer than 10 each and are omitted. \rev{The critical-path column lists ParaGUI and Claude Sonnet 4.6 mean steps side by side, not a ratio.}\label{tab:efficiency_partition_claude}}
    \centering
    \small
    \resizebox{\columnwidth}{!}{%
    \begin{tabular}{@{}llrcc c@{}}
      \toprule
      \textbf{Partition} & \rev{\textbf{Category}} & \textbf{\#} & \rev{\textbf{Critical path (Para vs.\ Claude)}} & \textbf{Step Red.\ ($S$)}~$\uparrow$ & \textbf{Token$\times$}~$\downarrow$ \\
      \midrule
      \multirow{2}{*}{Both Correct}
        & online shopping & 27 & 23.7\,/\,74.2 & \textbf{3.12$\times$} & \textbf{0.26$\times$} \\
        & web search      & 15 & 24.3\,/\,65.5 & 2.69$\times$          & 0.36$\times$ \\
      \midrule
      \multirow{3}{*}{Both Fail}
        & online shopping & 28 & 41.9\,/\,94.8 & 2.26$\times$ & 0.36$\times$ \\
        & web search      & 32 & 42.4\,/\,101.3 & 2.39$\times$ & 0.38$\times$ \\
        & file operation  & 27 & 54.3\,/\,74.4 & 1.37$\times$ & 0.46$\times$ \\
      \bottomrule
    \end{tabular}}%
\end{table}

\Cref{tab:efficiency_partition} (paired vs.\ Seed-1.8) highlights a clear category-level split in the \emph{both-correct} partition. On online shopping, ParaGUI shortens the critical path by $1.36\times$ at a small token overhead ($1.12\times$): parallel dispatch adds only a modest token cost when independent stores can be assigned to separate workers. On web search, the already-short Seed-1.8 trajectory leaves little room for step reduction; ParaGUI in fact uses about 22\% more steps ($S{=}0.82\times$); because self-contained dispatch (SCD) rewrites the full instruction and resolved dependencies into each worker's sub-task every round, the token cost inflates to $4.35\times$. In the \emph{both-fail} partition, step counts are roughly even---ParaGUI requires marginally more critical-path steps ($0.82$--$0.96\times$)---but token costs are uniformly $2.0$--$2.6\times$ higher. Because Seed-1.8 workers commit confidently to wrong answers (\Cref{tab:fail_step_dist}), the planner receives no signal that the trajectory has reached a dead end and continues dispatching parallel branches.

\Cref{tab:efficiency_partition_claude} (paired vs.\ Claude Sonnet 4.6) presents the complementary result. In both partitions, ParaGUI uses both fewer steps and fewer tokens: $2.7$--$3.1\times$ fewer steps with $0.26$--$0.36\times$ tokens on \emph{both-correct}, and $1.4$--$2.4\times$ fewer steps with $0.36$--$0.46\times$ tokens on \emph{both-fail}. The largest single-category gain is on both-correct online shopping ($3.12\times$ fewer steps at $0.26\times$ tokens). The two paired tables thus reveal a consistent pattern: against Seed-1.8 (the same worker backbone), ParaGUI trades extra tokens for higher success rate rather than for speed; against the token-heavier Claude Sonnet 4.6, it achieves both higher success rate and lower token usage.

\subsection{RQ2: When Parallel Execution Helps}
\label{sec:when_parallel}
Exploitable parallelism is governed primarily by task dependency structure rather than application domain alone: decomposable tasks reach a parallelism degree of approximately 2.2, whereas serial tasks remain near 1. Shared-document conflicts, already-short baseline trajectories, and dependence on visual context reduce the resulting efficiency gains.

\begin{figure*}[!t]
  \centering
  \includegraphics[width=\textwidth]{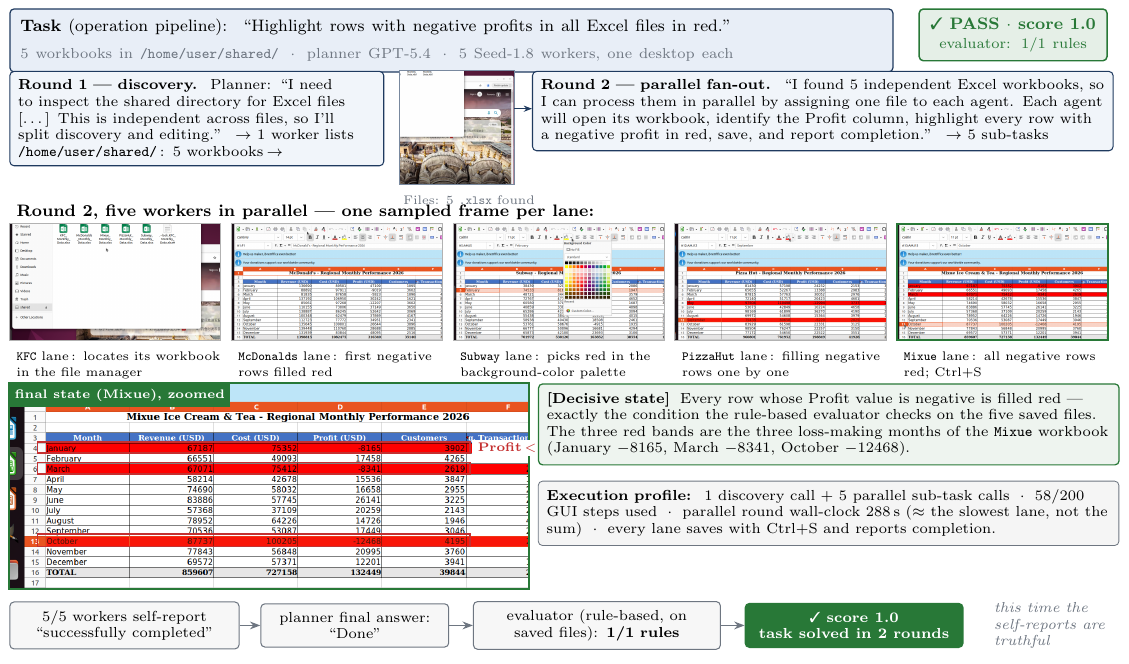}
  \caption{A successful ParaGUI trace on ParaGUIBench.
  \emph{Round 1}: one worker inspects the shared folder and reports five
  workbooks. \emph{Round 2}: the planner gives each of five workers one
  workbook; they run concurrently, so the round costs \rev{the most steps any single worker takes},
  not the sum. \emph{Zoom}: the final state---every negative-profit
  month filled red---is exactly what the rule-based evaluator checks
  (1/1 rules, score 1.0).}
  \label{fig:case_real_pass}
\end{figure*}

We measure exploitable parallelism by the parallelism degree $P = L_{\mathrm{total}}/L_{\mathrm{parallel}}$---how much shorter the critical path is than running the same decomposed sub-tasks serially. A task's intrinsic decomposability, \rev{rather than its application domain alone, primarily} governs $P$: the overall task-weighted degree is $P{=}2.13$, with per-category values in \Cref{tab:parallelism}.

\begin{table*}[!t]
\centering
\caption{Per-category parallelism on the full ParaGUIBench (233 tasks).
\emph{Worker steps} ($L_{\mathrm{total}}$) and \emph{Critical path} ($L_{\mathrm{parallel}}$) are per-task
averages; the latter matches the \emph{steps} column of \Cref{tab:main_results}. \emph{Para.}\ is the
parallelism degree $P=L_{\mathrm{total}}/L_{\mathrm{parallel}}$ (task-weighted). Bold marks the
per-column maximum.\label{tab:parallelism}}
\begin{tabular}{llrrrr}
\toprule
\textbf{Domain} & \rev{\textbf{Category}} & \textbf{\#} & \textbf{Worker steps (avg)} & \textbf{Critical path (avg)} & \textbf{Para.} \\
\midrule
\multirow{2}{*}{Info. Retrieval}
 & Web Search       & 65  & \textbf{108.0} & 38.9 & \textbf{2.78} \\
 & File Search      & 12  & 24.7  & 20.4 & 1.21 \\
\midrule
\multirow{4}{*}{Operation}
 & Online Shopping  & 91  & 71.4  & 32.7 & 2.18 \\
 & File Operation   & 42  & 92.1  & 54.5 & 1.69 \\
 & Web Navigate     & 13  & 47.2  & 29.2 & 1.61 \\
 & Search \& Write  & 10  & 87.1  & \textbf{59.2} & 1.47 \\
\midrule
\textbf{Total} &  & 233 & 82.2 & 38.7 & 2.13 \\
\bottomrule
\end{tabular}
\end{table*}

\Cref{fig:case_real_pass} traces a successful five-worker parallel run on one decomposable task. Grouping tasks by their dependency structure isolates where parallelism yields gains. \Cref{tab:by_parallel_class} reports SR, critical-path steps, and parallelism degree for the three parallel-pattern classes. The planner reduces parallelism to $P{\approx}1.1$ on \textit{serial} tasks and reaches $P{=}2.19$ and $P{=}2.28$ on the two decomposable (parallel-pattern) classes; this low serial value shows that round-based dispatch avoids over-dispatching when no parallelism is available. Meanwhile, SR remains similar across the three classes ($45.7$--$48.0\%$), suggesting that the higher parallelism on decomposable tasks is not associated with lower task success.

\begin{table*}[!t]
  \caption{ParaGUI results on ParaGUIBench (233 tasks) by parallel-pattern class. \textit{parallel\_independent}: sub-tasks with no inter-dependency; \textit{parallel\_dependent}: sub-task DAG; \textit{serial}: no exploitable parallelism. SR: Success Rate; Parallelism $=$ Worker steps \rev{$\div$} Critical-path steps.\label{tab:by_parallel_class}}
  \centering
  \begin{tabular}{@{}lrrrrrr@{}}
    \toprule
    \rev{\textbf{Parallel-pattern class}} & \textbf{\#} & \textbf{SR (\%)}
      & \textbf{Critical path (avg)} & \textbf{Worker steps (avg)}
      & \textbf{Para.}
      & \textbf{Tokens (M)} \\
    \midrule
    parallel\_independent & 164 & 45.73 & 33.97 &  74.45 & 2.19 & 0.73 \\
    parallel\_dependent   &  50 & 48.00 & 53.80 & 122.70 & 2.28 & 1.22 \\
    serial                &  19 & 47.37 & 39.58 &  43.05 & 1.09 & 0.43 \\
    \midrule
    \textbf{Total}        & 233 & 46.35 & 38.68 & 82.24
                          & 2.13 & 0.81 \\
    \bottomrule
  \end{tabular}
\end{table*}

The categories benefit through different mechanisms. The planner splits a web-search query into sub-queries that run on separate workers in parallel, and online shopping decomposes similarly by assigning each store to a different worker. File search instead benefits from giving each worker a single file, which removes the cross-file interference that degrades the single-agent baseline. The gains are eroded where parallelism cannot help: \emph{visual} web-search answers depend on page content that a text-only sub-task cannot convey, and search \& write stays hardest \rev{for both systems because workers must coordinate disjoint edits on one shared document; this contention also lengthens ParaGUI's critical path (59.2 vs.\ 48.0 steps} for Claude Sonnet 4.6).

\subsection{RQ3: Failure Modes}
\label{sec:failure}
\begin{figure*}[!t]
  \centering
  \includegraphics[width=\textwidth]{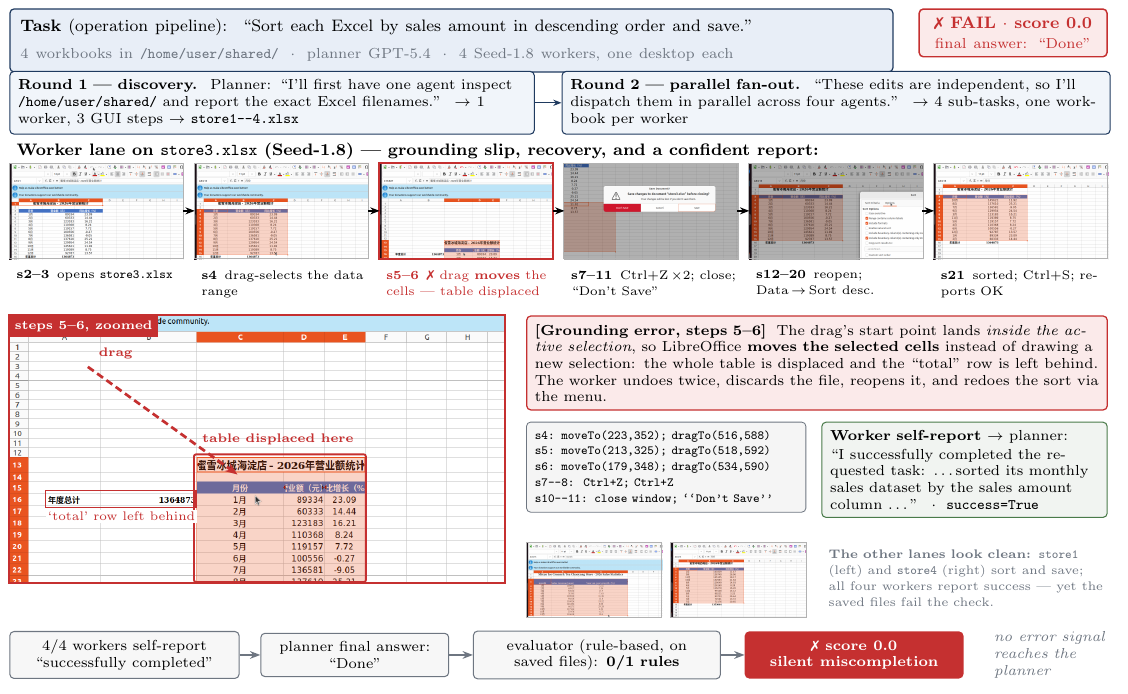}
  \caption{A silent miscompletion on ParaGUIBench. The planner decomposes
  correctly, and the \texttt{store3} worker even recovers from a grounding
  slip (a drag inside the active selection \emph{moves} the cells instead of
  selecting them; zoom). Yet all four workers report success, the planner
  answers ``Done'', and the rule-based evaluator rejects the saved files
  (0/1 rules, score 0.0)---no error signal ever reaches the planner.}
  \label{fig:case_real_fail}
\end{figure*}

Most ParaGUI failures are silent miscompletions in which the system reports completion even though the final environment state is incorrect. Among \rev{the 111 silent miscompletions, single-worker GUI execution is the largest root cause (53\% of the 111),} while planner decomposition and aggregation together account for 30\%.
\Cref{fig:case_real_fail} shows one such silent miscompletion: the worker recovers from a visual-grounding error during execution, yet the file it finally saves still fails the rule-based check. We now analyze the full failure set quantitatively.
We manually assign a primary root cause to each of ParaGUI's 125 failed tasks ($N{=}5$, full ParaGUIBench) by inspecting the execution trajectories case by case. The 125 failures fall into two groups: 111 silent miscompletions (89\%), where the system emits a substantive final answer and self-reports success while the true environment state is wrong, and 14 honest terminations (11\%) via a voluntary \texttt{fail()} call or global budget exhaustion. Explicit termination therefore accounts for only a small minority of failures.

We sort the 111 miscompletions into four root causes (shares of the 111 below). Single-worker GUI execution is the largest, at 59 cases (53\%): a worker gets an in-principle-solvable sub-task wrong (misread values; a search so literal that it reports an existing item as ``not found''; wrong visual judgment; mis-clicks). Planner errors form the second group (together 30\%): 20 \emph{decomposition} errors (18\%) and 13 \emph{aggregation} errors (12\%). A decomposition error partitions the task badly---for example, instructing each WebMall store to bookmark its own local cheapest item with no global comparison, or splitting a shared document so that workers overwrite each other (consistent with search \& write being the weakest category; \Cref{tab:by_parallel_class_cross}). An aggregation error instead merges correct per-worker results into a wrong final answer. The remaining 19 (17\%) are premature success claims, distinct from the 14 honest terminations above because here the worker wrongly reports success on an unsolved task.

The over-confident ``task-complete'' signal is therefore a symptom rather than a root cause in its own right: in decreasing order of frequency, it stems from single-worker GUI execution, planner decomposition and aggregation, and premature success claims.
It also sharpens how to read \Cref{tab:fail_step_dist}: most failures are confident wrong answers rather than early give-ups (honest \texttt{fail()} calls).

\Cref{tab:fail_step_dist} reports the average step count on failed tasks, broken down by \rev{category}. The main contrast is the gap between Seed-1.8 (36.6 avg.\ steps on fail) and Claude Sonnet 4.6 (80.1 avg.\ steps on fail). In Search \& Write, Seed-1.8 is no longer an extreme early-stop outlier: its 40.9 average steps are close to Claude's 47.1. Manual inspection of the underlying trajectories confirms that most early Seed-1.8 terminations are confident \texttt{done(answer)} actions that the evaluator rejects, rather than voluntary \texttt{fail()} calls; this is a \emph{calibration} failure---the same over-confident task-complete pattern identified above---not an early give-up. Claude Sonnet 4.6 appears less prone to early confident stopping: on tasks it cannot solve, it spends \emph{more} steps than its overall average (80.1 vs.\ 75.9) before failing. ParaGUI falls between the two baselines on failed tasks (42.8 steps): it inherits a Seed-1.8 worker but adds steps through planner-level re-dispatch. This re-dispatch gives additional opportunities for correction but does not verify the final environment state or detect silent miscompletions.

\begin{table}[!t]
  \centering
  \small
  \setlength{\tabcolsep}{5pt}
  \caption{Average critical-path steps on \emph{failed} tasks by \rev{\rev{category}, computed over each system's own set of failed tasks (the failed sets differ across systems, as the \#fail columns show).} Each cell: \#failed tasks and mean step count.\label{tab:fail_step_dist}}
  \begin{tabular}{@{}lcccccc@{}}
  \toprule
  \multirow{2}{*}{\rev{\textbf{Category}}} & \multicolumn{2}{c}{\textbf{Seed-1.8}} & \multicolumn{2}{c}{\textbf{ParaGUI}} &
  \multicolumn{2}{c}{\textbf{Claude 4.6}} \\
  \cmidrule(lr){2-3}\cmidrule(lr){4-5}\cmidrule(lr){6-7}
   & \textbf{\#fail} & \textbf{steps} & \textbf{\#fail} & \textbf{steps} & \textbf{\#fail} & \textbf{steps} \\
  \midrule
    Web Search & 49 & 39.1 & 38 & 42.9 & 44 & 96.5 \\
    File Search & 11 & 12.8 & 3 & 12.7 & 8 & 98.2 \\
    Online Shopping & 56 & 36.1 & 41 & 35.5 & 51 & 82.8 \\
    File Operation & 33 & 42.4 & 28 & 52.9 & 33 & 69.5 \\
    Web Navigate & 10 & 31.0 & 6 & 34.8 & 10 & 45.1 \\
    Search \& Write & 9 & 40.9 & 9 & 59.9 & 9 & 47.1 \\
  \midrule
    \textbf{Total} & \rev{168} & \rev{36.6} & \rev{125} & \rev{42.8} & \rev{155} & \rev{80.1} \\
  \bottomrule
  \end{tabular}
\end{table}

\subsection{RQ4: Component and Scaling Analysis}
\label{sec:ablation}

On the 51-task diagnostic subset, SCD is a major structural contributor: removing it lowers success from 39.2\% to 27.5\%. The worker backbone sets the capability ceiling, while increasing the worker count $N$ primarily shortens the critical path without monotonically improving success. This is consistent with the failure attribution above, where single-worker GUI execution was the largest root cause, although planner decomposition and aggregation still contribute a substantial share.
To quantify the contribution of each component, we conduct ablation studies on the same 51-task subset (on which ParaGUI scores 39.2\% SR, vs.\ 46.4\% on the full 233 tasks). \Cref{tab:ablation} summarizes the results. We provide the full parallelism scaling analysis in the appendix.

\paragraph{Component analysis}
The ablation isolates three findings. In \Cref{tab:ablation}, row labels follow a fixed convention: \texttt{plan-*} varies the planner while always using a Seed-1.8 worker, and \texttt{*-gui} varies the worker while always using a GPT-5.4 planner.

First, the gain is primarily associated with the planner's decomposition and SCD, not the planner model. With a Seed-1.8 worker, both a GPT-5.4 planner and a Seed-1.8 planner (\texttt{plan-seed-1.8}) reach 39.2\% success (a Kimi K2.5~\cite{cite5} planner, \texttt{plan-kimi}, reaches 33.3\%). All three far exceed the 17.6\% \rev{Seed-1.8 serial baseline (its worker backbone).} Disabling SCD ($-$SCD) drops the success rate to 27.5\%. Among these Seed-1.8-worker planner variants, the GPT-5.4 planner is the most efficient: it reaches the top success rate in the fewest critical-path steps (33.3 vs.\ 40.8 for \texttt{plan-seed-1.8}).

Second, \emph{the worker backbone sets the capability ceiling and governs efficiency}. A Claude Sonnet 4.6 worker (\texttt{claude-gui}) attains the highest SR (47.1\%); Seed-1.8 reaches 39.2\%, and Kimi K2.5 (\texttt{kimi-gui}) 35.3\%. Seed-1.8 remains the most efficient choice (33.3 steps and 0.73M tokens, vs.\ 50.0 steps and 1.98M tokens for Claude) and gives the best accuracy--efficiency trade-off among the three worker variants.

Third, \emph{$N$ shifts which tasks are solved without monotonically improving capability}. Across $N{=}1$, $3$, and $5$ on the 51-task subset, 18 tasks are solved by all three configurations; $N{=}1$ uniquely solves 3 further tasks, while $N{=}3$ and $N{=}5$ each unlock 2 tasks that $N{=}1$ misses. Two opposing mechanisms are at \rev{play. First, extra parallel branches introduce dispatch errors, so the $N{=}3$ and $N{=}5$ configurations lose the three tasks that only $N{=}1$ solves. Second, the same branching unlocks decompositions that a single worker cannot} complete within the 25-step per-invocation cap. Aggregate SR is therefore nearly constant and, if anything, slightly lower at larger $N$ (41.2\%, 39.2\%, and 39.2\% for $N{=}1$, $3$, $5$). Critical-path steps fall monotonically (60.5, 39.0, and 33.3, respectively), so the dominant effect of larger $N$ on this subset is a shorter critical path at nearly unchanged accuracy.

\begin{table}[!t]
  \caption{Ablation of the planner and worker on the 51-task subset.
  SR: overall Success Rate (All 51); steps: avg.\ critical-path steps;
  Tokens: avg.\ per-task tokens (M). Unless noted, parallel variants use a
  GPT-5.4 planner and Seed-1.8 worker; each variant's planner and worker pairing
  and the full per-category SR appear in \Cref{tab:ablation_full}.
  Rows \texttt{plan-*} vary the planner (worker fixed to Seed-1.8); rows \texttt{*-gui} vary the worker (planner fixed to GPT-5.4).
  \emph{$-$SCD} replaces SCD with a naive ``continue/finish'' \rev{dispatch.
  Bold marks the default ParaGUI configuration ($N{=}5$), not the per-column best. \rev{$\uparrow$/$\downarrow$: higher/lower is better.}\label{tab:ablation}}}
  \centering
  \small
  \setlength{\tabcolsep}{5pt}
  \begin{tabular}{@{}lccc@{}}
    \toprule
    \textbf{Configuration} & \textbf{SR (51)}~\rev{$\uparrow$} & \textbf{steps}~\rev{$\downarrow$} & \textbf{Tokens}~\rev{$\downarrow$} \\
    \midrule
    \multicolumn{4}{@{}l}{\textit{Serial baseline (GUI-only, no planner)}} \\
    \quad Seed-1.8                          & 17.6\% & 28.3 & 0.29 \\
    \quad Claude Sonnet 4.6                 & 35.3\% & 67.1 & 1.64 \\
    \midrule
    \multicolumn{4}{@{}l}{\textit{Parallel variants}} \\
    \quad plan-seed-1.8                     & 39.2\% & 40.8 & 0.74 \\
    \quad plan-kimi                         & 33.3\% & 45.6 & 0.88 \\
    \quad kimi-gui                          & 35.3\% & 54.0 & 1.37 \\
    \quad claude-gui                        & 47.1\% & 50.0 & 1.98 \\
    \quad ParaGUI $-$SCD ($N{=}5$)          & 27.5\% & 30.6 & 0.55 \\
    \quad ParaGUI ($N{=}1$)                 & 41.2\% & 60.5 & 0.63 \\
    \quad ParaGUI ($N{=}3$)                 & 39.2\% & 39.0 & 0.71 \\
    \quad \textbf{ParaGUI ($N{=}5$, def.)}  & \textbf{39.2\%} & \textbf{33.3} & \textbf{0.73} \\
    \bottomrule
  \end{tabular}
\end{table}

\subsection{RQ5: Transfer to OSWorld}
\label{sec:generalization}
On the full 369-task OSWorld suite, ParaGUI with $N{=}1$ improves overall SR from 51.5\% to 59.1\%, but its benefit is uneven: it \rev{matches the baseline on OS and Thunderbird and falls} behind on Chrome and GIMP, where tasks demand stronger visual continuity than text-only sub-tasks preserve.
To check whether our planner--worker architecture transfers beyond parallel-native tasks, we evaluate ParaGUI ($N{=}1$, serial) on the standard OSWorld benchmark~\cite{cite7} against Seed-1.8 in GUI-only mode. On the full 369-task suite, ParaGUI ($N{=}1$) reaches 59.1\%, ahead of the Seed-1.8 GUI-only baseline at 51.5\% ($+7.6$~points). This overall gain is not uniform: it reflects two opposing per-application effects (\Cref{tab:osworld}), with ParaGUI matching or outperforming Seed-1.8 on every application except Chrome and GIMP.

The dividing line is not single-application vs.\ multi-application, but whether a task can be split into self-contained textual sub-tasks without losing the visual context that each step depends on. Office and tool-operation tasks (Calc, Impress, VLC, VS~Code) decompose into bounded, independent steps that a stateless worker can execute from a one-line instruction (for example, ``enter the specified formula in A1'' or ``set the title font size to 18\,pt''); here the planner's global decomposition is a net gain over a single agent that accumulates context errors on long trajectories. Chrome and GIMP are instead continuous exploratory flows in which each step depends on prior visual state. Decomposing them into stateless, text-only fragments removes the visual continuity that subsequent actions need: the worker reads a textual log of what happened but cannot see the screens that produced it, and the planner, which receives no screenshots, must infer the missing state from text alone.

This gap is unlikely to be closed by changes to the planner or the text interface alone; the evidence points to persisting each worker's own visual frames across rounds as a likely missing factor. Overall, the planner--worker design transfers a net planning benefit to serial environments, conditional on how well each task can be expressed as text-only \rev{sub-tasks.}

Per application (\Cref{tab:osworld}), ParaGUI gains most on tool-operation and office tasks: LibreOffice Calc ($+27.6$~points), Impress ($+23.4$), VLC ($+17.7$), and VS~Code ($+13.0$). Gains are smaller on the multi-application split ($+8.9$) and Writer ($+4.4$). ParaGUI ties on OS and Thunderbird and falls behind only on Chrome ($-17.3$) and GIMP ($-15.4$). One Chrome case illustrates a likely cause: the round-2 worker received a 4{,}387-character verbatim history of the prior round yet still failed, suggesting that textual history alone is insufficient and that visual continuity is the missing factor---a target for future work.

\begin{table}[!t]
  \caption{OSWorld results per application. ParaGUI \rev{($N{=}1$)---a GPT-5.4 planner coordinating a single Seed-1.8 worker (no parallel worker execution)---vs.\ the Seed-1.8 GUI-only baseline} on the full 369-task suite. Missing results counted as 0; SR is binary (success only if OSWorld score $= 1.0$). \rev{$\uparrow$: higher is better.}\label{tab:osworld}}
  \centering
  \small
  \resizebox{\columnwidth}{!}{%
  \begin{tabular}{@{}lrcrcr@{}}
    \toprule
    & & \textbf{ParaGUI} ($N{=}1$) && \textbf{GUI-only} (Seed-1.8) \\
    \cmidrule(lr){3-3} \cmidrule(lr){5-5}
    \textbf{Application} & \textbf{\#} & \textbf{SR}~\rev{$\uparrow$} && \textbf{SR}~\rev{$\uparrow$} \\
    \midrule
    chrome              &  46 &  45.7\% &&  63.0\% \\
    gimp                &  26 &  65.4\% &&  80.8\% \\
    libreoffice\_calc   &  47 &  78.7\% &&  51.1\% \\
    libreoffice\_impress&  47 &  68.1\% &&  44.7\% \\
    libreoffice\_writer &  23 &  60.9\% &&  56.5\% \\
    multi\_apps         & 101 &  30.7\% &&  21.8\% \\
    os                  &  24 &  75.0\% &&  75.0\% \\
    thunderbird         &  15 &  86.7\% &&  86.7\% \\
    vlc                 &  17 &  82.4\% &&  64.7\% \\
    vs\_code            &  23 &  91.3\% &&  78.3\% \\
    \midrule
    \textbf{Overall}    & \textbf{369} & \textbf{59.1\%} && \textbf{51.5\%} \\
    \bottomrule
  \end{tabular}}%
\end{table}

\section{Conclusion}
\label{sec:conclusion}

We identified the sequential interaction paradigm as a key efficiency bottleneck for GUI agents on long-horizon complex tasks. We took a first step toward evaluating multi-agent parallel \rev{coordination} in this setting. We introduced ParaGUIBench, a 233-task benchmark for parallel execution and coordination in realistic stateful GUI environments. It is backed by a parallel-native multi-device Docker infrastructure and efficiency-oriented metrics. We also introduced ParaGUI, a planner--worker agent that exploits the parallel tool-calling capability of modern LMMs. ParaGUI reaches up to $2.78\times$ parallelism on web search, with an overall task-weighted value of $2.13\times$ across ParaGUIBench. On the full 233-task benchmark, it raises success rate from 33.5\% (the strongest serial baseline, Claude Sonnet 4.6) to 46.4\% at roughly half the critical-path steps and less than half the tokens. When run in a serial (single-worker) configuration, it even outperforms a strong GUI-only baseline on the standard OSWorld benchmark (59.1\% vs.\ 51.5\%), with its only structural shortfall on the applications that demand the most visual continuity, Chrome and GIMP.

\paragraph{Limitations}
Our analysis surfaces four limitations. First, \emph{the worker backbone, not the planner, is the dominant bottleneck}. Most ParaGUI failures are confident-but-wrong completions whose leading cause is single-worker GUI execution rather than running out of steps; because the planner sees only an over-confident success signal (\Cref{tab:fail_step_dist}), it cannot detect a dead-end trajectory and keeps dispatching. These undetected worker failures both cap capability and amplify the token overhead vs.\ the token-efficient Seed-1.8 baseline ($2.1\times$ on the full 233-task benchmark, though still $0.4\times$ vs.\ the token-heavier Claude Sonnet 4.6). Second, the planner delegates disjoint-region assignment entirely to in-context reasoning, so current LMMs struggle to keep sub-tasks disjoint when they share a document, yielding decomposition and aggregation errors. Third, the synchronous round barrier makes ParaGUI sensitive to uneven sub-task lengths, because the planner cannot start the next round until every worker in the current round returns. Fourth, our experiments cover a limited set of LMM backbones: GPT-5.4 on the planner side; Seed-1.8, Qwen3-VL-235B-A22B-Instruct, Holo3-35B-A3B, Claude Sonnet 4.6, and Kimi K2.5 as workers or baselines. We leave broader coverage of worker and planner backbones to future work.

\paragraph{Future Work}
These limitations point to several concrete directions. A natural next step is \emph{adaptive task allocation}: a gating mechanism that switches between parallel and serial execution based on task complexity. This is most promising on information-retrieval tasks, which exhibit the largest parallelism headroom (web search reaches $P{=}2.78$; \Cref{tab:parallelism}). \emph{Worker-side calibration} training (e.g.\ reinforcement-learning (RL)-trained abstention) directly targets the over-confident commitment shown in \Cref{tab:fail_step_dist}, \rev{a major driver of} ParaGUI's token overhead vs.\ Seed-1.8 GUI-only; planner-side RL over the round-based dispatch interface targets disjoint-region assignment and earlier termination of dead-end branches. Our OSWorld analysis points to a further direction: letting each worker persist its own visual frames---not merely the planner's textual relay---across rounds, which is \rev{one promising} way to recover the visual continuity that text-mediated dispatch loses on vision-heavy applications such as Chrome and GIMP. Action-conditioned world models studied in other visual decision domains~\cite{ren2025surfer} suggest one possible route to maintaining such visual state across dispatch rounds, although adapting them to GUI environments remains open. More broadly, allowing workers to exchange more than file \rev{artifacts}---for example, through structured messaging or shared memory---would extend the parallel paradigm beyond storage-only synchronization.
Asynchronous dispatch and adaptive load balancing are natural extensions that directly target the synchronous round barrier noted above.

\bibliographystyle{IEEEtran}
\bibliography{main}

\appendices

\section{Extended Experimental Analysis}
\label{sec:appendix_analysis}

\subsection{Efficiency by Parallel-Execution Pattern}
\label{sec:appendix_analysis_pattern}

A breakdown by parallel class $\times$ \rev{category} (\Cref{tab:by_parallel_class_cross}) shows where the largest parallelism gains accrue. The biggest gains come from web search and online shopping within \textit{parallel\_independent} and \textit{parallel\_dependent}: in these \rev{categories}, each sub-task admits a separate browser context. The search\,\&\,write category is limited by a shared-document bottleneck: its sub-tasks edit the same file, so parallelism reaches only $1.47$ and SR drops to $10\%$. Concurrent edits on one document expose a limitation of in-context planning---it cannot reliably keep workers' edits within disjoint regions.

\begin{table}[!t]
  \caption{ParaGUI on ParaGUIBench broken down by parallel class $\times$ \rev{category}
    (cells with $n\!\leq\!2$ are reported but should be interpreted with caution).
    The parallel class is a task's annotated intrinsic dependency structure; the actually dispatched parallelism degree can still exceed~1 when the planner over-dispatches, so a few \textit{serial}-class cells (e.g.\ Web Navigate, $n{=}2$) report $P{>}1$.\label{tab:by_parallel_class_cross}}
  \centering
  \resizebox{\linewidth}{!}{%
  \begin{tabular}{@{}llrrrrr@{}}
    \toprule
    \rev{\textbf{Parallel-pattern class}} & \rev{\textbf{Category}} & \textbf{\#}
      & \textbf{SR (\%)} & \rev{\textbf{Critical path} (avg)}
      & \rev{\textbf{Para.}} & \textbf{Tokens (M)} \\
    \midrule
    \multirow{6}{*}{parallel\_independent}
      & Web Search      & 30 & 36.67 & 31.60 & 2.62 & 0.83 \\
      & File Search     &  8 & 87.50 & 20.25 & 1.20 & 0.21 \\
      & Online Shopping & 75 & 52.00 & 28.49 & 2.51 & 0.69 \\
      & File Operation  & 30 & 33.33 & 47.27 & 1.98 & 0.96 \\
      & Web Navigate    & 11 & 63.64 & 28.55 & 1.59 & 0.43 \\
      & Search \& Write & 10 & 10.00 & 59.20 & 1.47 & 0.87 \\
    \midrule
    \multirow{3}{*}{parallel\_dependent}
      & Web Search      & 32 & 46.88 & 47.44 & 2.95 & 1.38 \\
      & Online Shopping &  8 & 75.00 & 57.50 & 1.64 & 0.93 \\
      & File Operation  & 10 & 30.00 & 71.20 & 1.26 & 0.96 \\
    \midrule
    \multirow{5}{*}{serial}
      & Web Search      &  3 & 33.33 & 20.67 & 1.00 & 0.19 \\
      & File Search     &  4 & 50.00 & 20.75 & 1.23 & 0.26 \\
      & Online Shopping &  8 & 62.50 & 47.50 & 1.00 & 0.49 \\
      & File Operation  &  2 & 50.00 & 80.50 & 1.00 & 0.79 \\
      & Web Navigate    &  2 &  0.00 & 33.00 & 1.71 & 0.55 \\
    \bottomrule
  \end{tabular}}%
\end{table}

\subsection{Extended Ablation Discussion}
\label{sec:appendix_analysis_ablation}

\begin{figure*}[!t]
  \centering
  \includegraphics[width=0.92\linewidth]{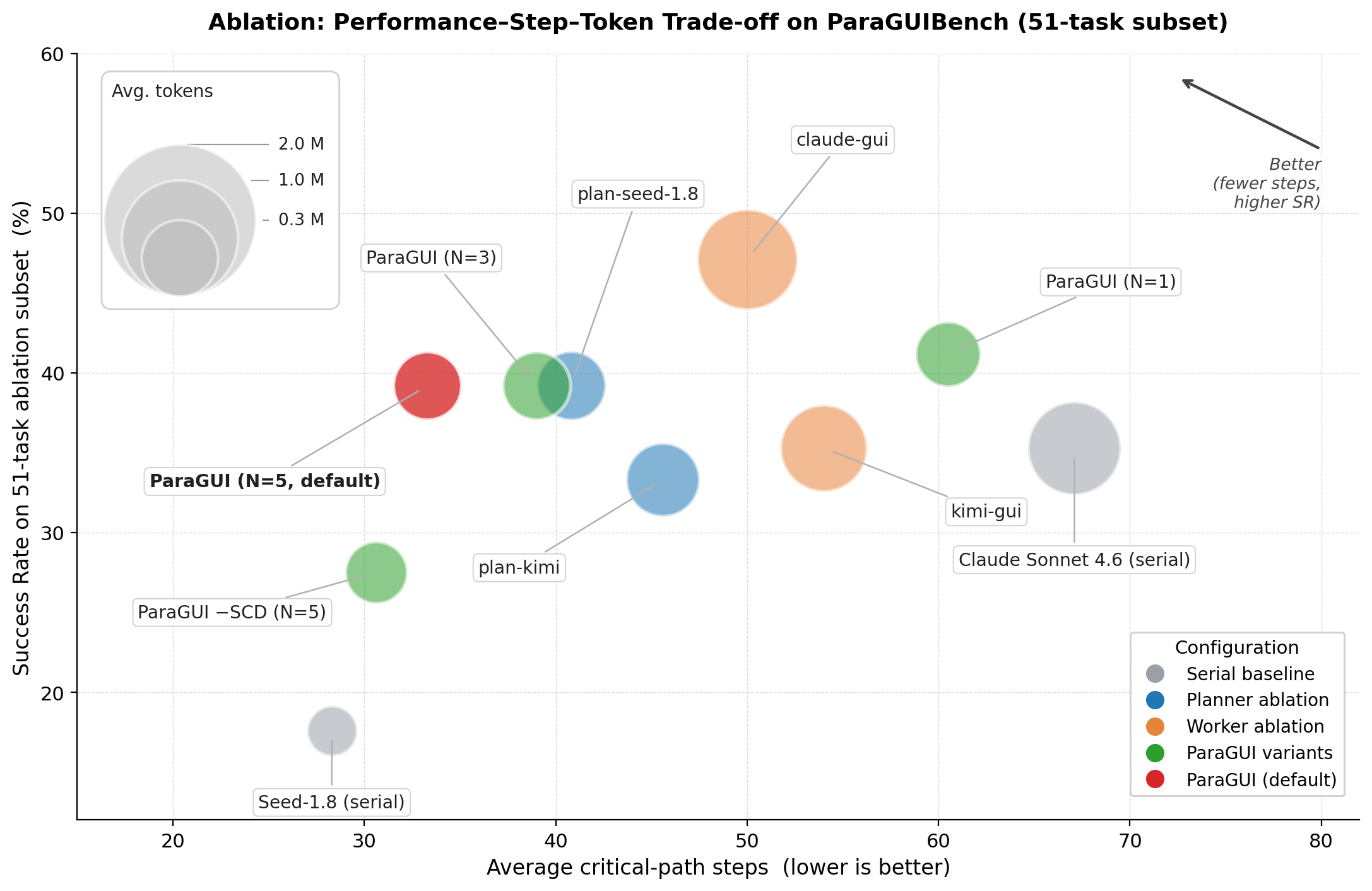}
  \caption{Ablation on the 51-task subset in the capability--efficiency plane.
  Vertical: Success Rate ($\uparrow$); horizontal: avg.\ critical-path steps
  ($\downarrow$); bubble area $\propto$ avg.\ per-task tokens. Colour groups the
  comparison type (serial baseline / planner / worker / $-$SCD / $N$-sweep); the
  dotted path traces $N{=}1\to3\to5$, where steps shrink at near-constant SR.
  ParaGUI ($N{=}5$, default) is the fastest high-SR Seed-1.8-worker
  variant, while $N{=}1$ obtains slightly higher SR at substantially more
  steps. Numbers in \Cref{tab:ablation_full}.}
  \label{fig:ablation_bubble}
\end{figure*}

\Cref{tab:ablation_full} reports the full per-category breakdown of the ablation summarized in \Cref{tab:ablation}, and \Cref{fig:ablation_bubble} visualizes the same comparison in the capability--efficiency plane.

The 51-task subset approximates the full benchmark's domain mix (IR 33.3\%, Shop 39.2\%, FileOps 17.6\%, WebNav 3.9\%, SearchWrite 5.9\%). Relative to the Seed-1.8 serial baseline (17.6\%, 28.3 steps, 0.29M tokens), ParaGUI (39.2\%, 33.3 steps, 0.73M tokens) lifts SR by +21.6~points while spending a few extra steps and roughly \rev{$2.5\times$} more tokens. The step parity, rather than a step reduction, traces to a baseline behavior pattern. Seed-1.8 frequently commits to a confident-but-wrong answer well before the step budget: averaged over failed tasks, Seed-1.8 spends 36.6 steps versus 80.1 for the well-calibrated Claude Sonnet 4.6 (\Cref{tab:fail_step_dist}), with the \rev{search} \& write row at 40.9 versus 47.1. Almost none of these early terminations are voluntary \texttt{fail()} calls; they are confident \texttt{done(answer)} that the evaluator rejects. This pattern suppresses Seed-1.8's step and token counts without improving SR. ParaGUI continues dispatching through additional rounds, lifting failed-task step counts to 42.8 on average---closer to well-calibrated behavior despite using a Seed-1.8 worker (\Cref{tab:fail_step_dist}); these extra rounds give the planner more chances to correct a worker's output, though they do not verify the final environment state. The clean step reduction visible on the full benchmark (38.7 vs.\ 75.9) appears precisely once the comparison baseline switches to Claude Sonnet 4.6, which does \emph{not} exhibit early confident-but-wrong commitment. Isolating SCD directly, the $-$SCD variant reaches only 27.5\% versus 39.2\% for the full system; this confirms SCD as the main capability lever.

Holding the worker fixed to Seed-1.8, we vary the planner. A Seed-1.8 planner (plan-seed-1.8) matches the GPT-5.4 planner in overall success (39.2\%). A Kimi K2.5 planner (plan-kimi) trails only modestly (33.3\%). Both remain far above the 17.6\% serial baseline. The planner model is therefore \emph{not} the capability bottleneck on this subset. Once decomposition and SCD are in place, even a Seed-1.8 planner produces sub-tasks the worker can execute. The GPT-5.4 planner helps efficiency: it reaches 39.2\% in 33.3 critical-path steps, against 40.8 (plan-seed-1.8) and 45.6 (plan-kimi). These are tighter, less redundant decompositions.

We next fix the planner to GPT-5.4 and vary the worker. A Claude Sonnet 4.6 worker (claude-gui) attains the highest SR (47.1\%), followed by Seed-1.8 (39.2\%) and Kimi K2.5 (kimi-gui, 35.3\%). All three more than double the 17.6\% serial baseline. Seed-1.8 is the most efficient (33.3 steps, 0.73M tokens, versus 54.0/1.37M for Kimi and 50.0/1.98M for Claude). It offers the best accuracy--efficiency trade-off. Claude is the strongest worker when capability matters more than cost.

Varying the parallelism degree shifts \emph{which} tasks are solved without monotonically improving aggregate capability. Across $N{=}1/3/5$, the three configurations \rev{all solve the same 18 tasks}; $N{=}1$ uniquely solves 3 more tasks that $N{=}3/5$ miss, while $N{=}3$ and $N{=}5$ each unlock 2 tasks that $N{=}1$ misses. Two opposing mechanisms are visible. (i) Extra parallel branches incur additional dispatch errors that cost $N{=}3/5$ the three $N{=}1$-only tasks: the planner has to split a coherent trajectory across workers, and any single dispatch boundary error fails the whole task. (ii) The same parallelism unlocks decompositions that a single worker cannot complete within the 25-step per-invocation cap: an under-budgeted serial trajectory simply runs out of room. The two effects nearly cancel on this subset, leaving aggregate SR essentially flat (41.2 / 39.2 / 39.2). What $N$ changes monotonically is \emph{efficiency}: the critical-path step count falls $60.5 \to 39.0 \to 33.3$. The planner--worker pair sets the capability ceiling; \rev{larger $N$ gives up little accuracy for a substantially shorter critical path.}

\subsection{Visual-History Budget: Why ParaGUI Saves Tokens vs.\ Claude Sonnet 4.6}
\label{sec:appendix_analysis_visual_budget}

A natural concern is whether ParaGUI's token advantage over Claude Sonnet 4.6 (\Cref{tab:efficiency_partition_claude}) is an \rev{artifact} of the visual-history configuration: ParaGUI's Seed-1.8 worker retains 3 screenshots, while Claude's default policy retains up to 10. We test this directly by restricting Claude Sonnet 4.6 to a 3-screenshot budget (matching Seed-1.8's policy) on the 51-task subset; the result is the row labelled ``GUI-only (image=3)'' in \Cref{tab:ablation_full}. The restriction does \emph{not} reduce Claude's token cost. Overall SR drops slightly (35.3\% $\to$ 33.3\%), critical-path steps \emph{increase} (67.1 $\to$ 78.0), and total tokens \emph{increase} (1.64M $\to$ 1.76M): with fewer screenshots in context the agent compensates by re-grounding for more steps, and the extra steps more than offset the per-step token savings. Under this normalized visual-history budget, ParaGUI's token use relative to Claude drops from $0.45\times$ to \rev{$0.42\times$ on this 51-task subset ($0.73$M${}/{}1.64$M vs.\ $0.73$M${}/{}1.76$M; \Cref{tab:ablation_full}),} widening its token advantage. This rules out a pure visual-context-budget account of the token gap and indicates that Claude's default 10-screenshot retention is closer to its own best operating point on these tasks. The token gap reported in the main results therefore reflects intrinsic per-call cost and trajectory length, not under-allocated visual context.

\begin{table*}[t]
\centering
\caption{Full per-category ablation on the 51-task subset, with Information Retrieval (IR)
split into \textbf{File Search} and \textbf{Web Search}. SR, steps, Tokens follow \Cref{tab:ablation}.
``---'': no planner (serial). ParaGUI ($N{=}5$): default system; $-$SCD: naive dispatch,
isolating SCD's contribution. Web Search includes the single VisualSearch task, following the
main-table taxonomy.}
\label{tab:ablation_full}
\resizebox{\textwidth}{!}{%
\begin{tabular}{lll ccccccc cc}
\toprule
 & & & \multicolumn{7}{c}{\textbf{Success Rate}} & & \\
\cmidrule(lr){4-10}
\textbf{Configuration} & \textbf{planner} & \textbf{worker}
& \textbf{File Search (7)} & \textbf{Web Search (10)} & \textbf{Shop (20)} & \textbf{WebNav (2)}
& \textbf{FileOps (9)} & \textbf{SearchWrite (3)} & \textbf{All (51)} & \textbf{steps} & \textbf{Tokens (M)} \\
\midrule
\multicolumn{12}{l}{\textit{Serial baseline}} \\
GUI-only            & ---       & Seed-1.8           & 0.0\%  & 30.0\% & 20.0\% & 100.0\% & 0.0\%  & 0.0\% & 17.6\% & 28.3 & 0.29 \\
GUI-only            & ---       & Claude Sonnet 4.6  & 28.6\% & 30.0\% & 40.0\% & 100.0\% & 33.3\% & 0.0\% & 35.3\% & 67.1 & 1.64 \\
GUI-only (image=3)  & ---       & Claude Sonnet 4.6  & 57.1\% & 10.0\% & 50.0\% & 100.0\% & 0.0\%  & 0.0\% & 33.3\% & 78.0 & 1.76 \\
\midrule
\multicolumn{12}{l}{\textit{Parallel variants (ablating planner or worker)}} \\
plan-seed-1.8                       & Seed-1.8  & Seed-1.8          & 71.4\% & 30.0\% & 35.0\% & 50.0\%  & 44.4\% & 0.0\% & 39.2\% & 40.8 & 0.74 \\
plan-kimi                           & Kimi K2.5 & Seed-1.8          & 57.1\% & 40.0\% & 30.0\% & 50.0\%  & 22.2\% & 0.0\% & 33.3\% & 45.6 & 0.88 \\
kimi-gui                            & GPT-5.4   & Kimi K2.5         & 71.4\% & 30.0\% & 30.0\% & 100.0\% & 22.2\% & 0.0\% & 35.3\% & 54.0 & 1.37 \\
claude-gui                          & GPT-5.4   & Claude Sonnet 4.6 & 71.4\% & 50.0\% & 50.0\% & 100.0\% & 22.2\% & 0.0\% & 47.1\% & 50.0 & 1.98 \\
ParaGUI $-$SCD ($N{=}5$)            & GPT-5.4   & Seed-1.8          & 71.4\% & 20.0\% & 25.0\% & 100.0\% & 0.0\%  & 0.0\% & 27.5\% & 30.6 & 0.55 \\
ParaGUI ($N{=}1$)                   & GPT-5.4   & Seed-1.8          & 85.7\% & 30.0\% & 40.0\% & 100.0\% & 22.2\% & 0.0\% & 41.2\% & 60.5 & 0.63 \\
ParaGUI ($N{=}3$)                   & GPT-5.4   & Seed-1.8          & 71.4\% & 30.0\% & 35.0\% & 100.0\% & 33.3\% & 0.0\% & 39.2\% & 39.0 & 0.71 \\
\textbf{ParaGUI ($N{=}5$, default)} & \textbf{GPT-5.4} & \textbf{Seed-1.8} & \textbf{71.4\%} & \textbf{30.0\%} & \textbf{40.0\%} & \textbf{100.0\%} & \textbf{22.2\%} & \textbf{0.0\%} & \textbf{39.2\%} & \textbf{33.3} & \textbf{0.73} \\
\bottomrule
\end{tabular}%
}
\end{table*}

\end{document}